# Object Disparity

Ynjiun Paul Wang

paulwang@microai.systems

August 17, 2021
Cupertino, CA

## Abstract

Most of stereo vision works are focusing on computing the dense pixel disparity of a given pair of left and right images. A camera pair usually required lens undistortion and stereo calibration to provide an undistorted epipolar line calibrated image pair for accurate dense pixel disparity computation. Due to noise, object occlusion, repetitive or lack of texture and limitation of matching algorithms, the pixel disparity accuracy usually suffers the most at those object boundary areas. Although statistically the total number of pixel disparity errors might be low (under 2% according to the Kitti Vision Benchmark of current top ranking algorithms), the percentage of these disparity errors at object boundaries are very high. This renders the subsequence 3D object distance detection with much lower accuracy than desired. This paper proposed a different approach for solving a 3D object distance detection by detecting object disparity directly without going through a dense pixel disparity computation. An example squeezenet Object Disparity-SSD (OD-SSD) was constructed to demonstrate an efficient object disparity detection with comparable accuracy compared with Kitti dataset pixel disparity ground truth. Further training and testing results with mixed image dataset captured by several different stereo systems may suggest that an OD-SSD might be agnostic to stereo system parameters such as a baseline, FOV, lens distortion, even left/right camera epipolar line misalignment.

## 1. Introduction

Stereo vision is one of the crucial vision functions to enable humans to drive a car with vision depth cues. Although physiologically one can show that a human's eye-brain is capable of resolving pixel level disparity[1], the object level disparity (distance/velocity) detection seems to be more relevant to a driving scene understanding rather than pixel level disparity. It has been demonstrated by [2] that multiple non-stereo calibrated cameras using recurrent neural networks are capable of object depth sensing even without a Lidar or Radar.

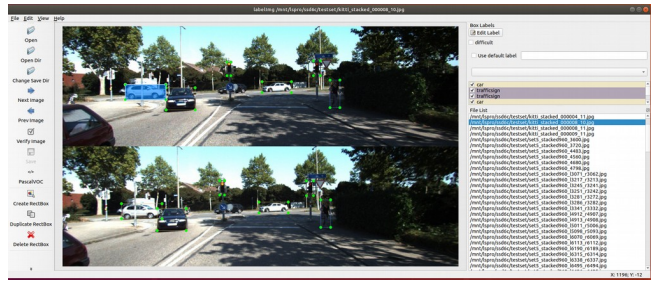

*Figure 1: labelImgPair tool for annotating object disparity*

There are several limitations in computing pixel level (dense) disparity: 1) noise sensitivity; 2) no depth information at an object occlusion area; 3) not reliable at lack of texture areas or repetitive texture areas; 4) local vs. global matching cost optimization competing criteria dilemma. The root cause of these limitations is that the pixel is a very weak (not reliable) matching target for computing disparity regardless in traditional stereo matching approach[3][4] or in deep learning approach[5][6] . To fundamentally address this issue, this paper proposed to use a stronger and higher level feature matching for disparity computation – namely, object level disparity.

In this paper, we start by defining what is object disparity in section 2. Then introduce a labeling tool for annotating an object disparity in a stacked stereo image pair. In section 3, an example squeezenet Object Disparity-SSD (OD-SSD) is constructed, then trained by a stacked stereo dataset converted from a 2015 Kitti stereo dataset. In section 4, a 95% percentile disparity in a predicted bounding box is defined over Kitti disparity map ground truth for testing OD-SSD accuracy against Kitti ground truth. In section 5, we benchmarked OD-SSD speed by running at several edge computing devices. In section 6, we demonstrated the robustness and stereo-parameter-agnostic of OD-SSD by using a mixed dataset captured by several different uncalibrated stereo system with different parameters for training and testing.

## 2. Object Disparity

An object disparity *dx, dy* is defined as below:

> Let an object in a left image with bounding box pixel coordinate as [*lxmin,lymin,lxmax,lymax*] and the same object in a right image with bounding box coordinate as [*rxmin,rymin,rxmax,rymax*], then the object disparity is
>
> *if lxmin==0 or rxmin==0:*
>     *dx=lxmax-rxmax*
> *elif lxmax==img_width or rxmax==img_width:*
>     *dx=lxmin-rxmin*
> *else*
>     *dx=(lxmin+lxmax)/2-(rxmin+rxmax)/2*
>
> *if lymin==0 or rymin==0:*
>     *dy=lymax-rymax*
> *elif lymax==img_height or rymax==img_height:*
>     *dy=lymin-rymin*
> *else*
>     *dy=(lymin+lymax)/2-(rymin+rymax)/2*

Conceptually an object disparity is defined as the same as a pixel disparity when in an extreme case an object size is reduced to a pixel.

For a stereo epipolar line calibrated left/right image pair, the *dy* will be zero. Interestingly, as we can see in a moment, the OD-SSD training and testing dataset does not require a stereo epipolar line calibrated left/right image pair. Thus *dy* may not always be zero yet still maintain the accuracy of *dx*. This nice behaviour relaxes a stereo camera calibration requirement thus reducing the stereo camera system complexity and the cost significantly.

For annotating object disparity, we modified a labeling tool [7] , and called it labelImgPair, and made it accept a stacked stereo image pair (or a stereo image). A stacked image pair is made of a left image on the top and a right image at the bottom as shown in Figure 1.

An extended Pascal VOC xml annotation file is generated by the labelImgPair as shown in Figure 2.

*Figure 2: Object Disparity annotation in an extended Pascal VOC format*

An object consists of two bounding boxes: a left image bndbox and a right image bndbox2. The disparity *dx, dy* is wrapped in the <delta> element.

## 3. Object Disparity-SSD

Recent success on 2D object detection using SSD [8] motivates us to extend its 2D architecture into OD-SSD which is a 3D architecture in nature capable of detecting a stereo pair of bounding boxes of an object and its disparity. After several iterations of design phase training and testing, we settled down at the following architecture:

Feature Map extraction base net uses the sequeezenet version 1.1 [9] with last two Fire layers output channels modification for fitting OD-SSD classification and regression heads as shown in Figure 3.

*Figure 3: squeeznet 1.1 with last two layers modification*

OD-SSD folds the stacked 2D (left at top half and right at bottom half) feature map into a single "3D" feature map with concatenated channels. For example, the output of the base net layer 11 is a stacked 2D feature map [3, 256, 40, 40]. After folding, the 3D feature map become [3, 512, 20, 40]. The Figure 4 shows the OD-SSD layers. As noted, the regression_headers output channels are changed from 6*4 (where 4 dimensions are bounding box coordinates: *cx, cy, w, h*) to 6*6 (where 6 dimensions are bounding box coordinates plus object disparity: *cx, cy, w, h, dx, dy*).

This proposed OD-SSD has fundamentally changed the disparity problem from a searching pixel level disparity as a stereo *matching* problem into a detecting object disparity as an object *detection* problem.

*Figure 4: OD-SSD*

## 4. Training and Testing

The 2015 KITTI Vision Benchmark Suite – Stereo [10]– is used for training and testing. There are 200 training scenes and 200 test scenes (4 color images per scene). Since OD-SSD does not require dense pixel disparity ground truth, we reverse the role of testing dataset as our training dataset and manually annotate the object disparity bounding boxes using labelImgPair tool as our ground truth for training. Another reason we reverse the role of training and testing dataset is we would like to compare our object disparity testing results with dense pixel disparity ground truth using $95^{th}$ percentile disparity value within the predicted object bounding box.

We selected 514 stereo image pairs as training dataset (200 test scenes which is 400 stereo image pairs plus 57 training scenes which is 114 stereo image pairs) and 86 stereo image pairs as testing dataset (43 training scenes which is 86 stereo image pairs).

In order to recognize vehicles from far away, we constructed our model to accept an input left/right image resolution of 640x320. By stacking the left image at the top and the right image at the bottom, the input stereo image resolution thus becomes 640x640. After training for 635 epochs with batch size of 44, we start our testing.

Let's use one of 86 test stereo images shown in the Figure 5 to show case what a test result looks like.

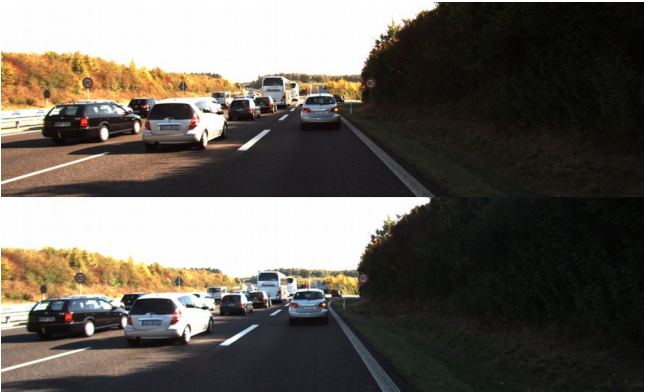

*Figure 5: a test stereo image (top: left image and bottom: right image)*

The test result of above stereo image is shown in the Figure 6. The top image is the left image with bounding boxes reporting the category recognized and the confidence level. The bottom image is the right image with bounding boxes reporting predicted object disparity (top) and object disparity ground truth (bottom). For example, the leftmost black car has the predicted object disparity of dx=33, dy=1 and the object disparity ground truth of dx=33, dy=1. Similarly, the rightmost white car has the predicted object disparity of dx=25, dy=0 and the object disparity ground truth of dx=24, dy=0. As we know, that Kitti dataset is a lens undistorted and stereo calibrated dataset. Thus the ground truth dy=0 is consistent with the dataset spec.

To compare the testing result against the Kitti dense disparity map ground truth, we define a bounding box 95$^{th}$ percentile disparity as below:

$$disparity_{bbox}(95\%) = disparity\ value\ at\ 95th\ percentile \in bbox$$

In Figure 7, the corresponding Kitti disparity map ground truth with 95$^{th}$ percentile disparity value within the predicted bounding box (Kitti 95% dx) is shown.

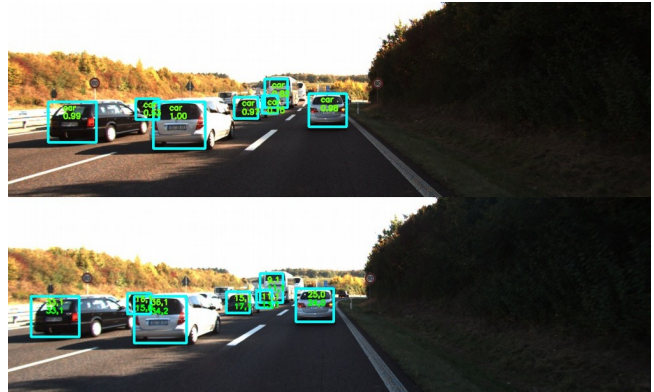

*Figure 6: OD-SSD test result at 640x640 input resolution*

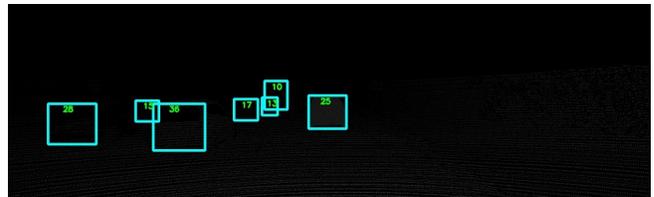

*Figure 7: corresponding Kitti disparity map ground truth with 95$^{th}$ precentile disparity (Kitti 95% dx) in each predicted bounding box*

The following table lists the Figure 6 and Figure 7 teseting results vs. ground truth from left to right bounding boxes.

| Table 1: Figure 5 stereo image OD-SSD testing result | | | | | | |
|---|---|---|---|---|---|---|
| Bbox (from left to right) | Obj detect-ed | Confid-ence | Predict-ed dx | Ground truth dx | Kitti 95% dx | Kitti 97% dx |
| 1 | car | 0.99 | 33 | 33 | 28 | 31 |
| 2 | car | 0.53 | 16 | 15 | 15 | 15 |
| 3 | car | 1.00 | 36 | 34 | 36 | 36 |
| 4 | car | 0.97 | 15 | 17 | 17 | 17 |
| 5 | car | 0.75 | 11 | 13 | 13 | 16 |
| 6 | car | 0.82 | 9 | 11 | 10 | 10 |
| 7 | car | 0.98 | 25 | 24 | 25 | 25 |

Interesting observation is that if we compared against Kitti 95% dx, the predicted dx is withing 2 pixels of error except for the leftmost black car (bbox #1), which has a disparity error of 5 pixels. However, if we compared against Kitti 97% dx, then the disparity error becomes 2 pixels for bbox#1. The reason for such phenomena is due to the perspective view of the bbox#1. The object disparity in this case predicts the closest distance of the leftmost black car, which is more relevant to autonomous driving applications. Bbox#5 is another interesting example of a partially occluded car. For bbox#5 Kitti 97% dx, it might take the dx value of the car in front (bbox#4) thus rendering inaccurate disparity estimate of dx=16 of bbox#5 car. In contrast, the bbox#5 car predicted dx = 11 is closer to the reality of dx=13 voted by both ground truth dx and Kitti 95% dx.

The overall test results of 86 stereo images test dataset can be summarized in Table 2 below.

| Table 2: OD-SSD test results at 640x640 | |
|---|---|
| Total test stereo images | 86 |
| Precision (car*) | 91.4% |
| Recall (car*) | 77.1% |
| Mean abs obj disparity error | 1.62 pixels |
| Max abs obj disparity error | 6.87 pixels |
| Abs obj disparity error histogram (pixel) | Number of objects |
| [0, 1) | 117 |
| [1, 2) | 64 |
| [2, 3) | 48 |
| [3, 4) | 19 |
| [4, 5) | 10 |
| [5, 6) | 6 |
| [6, 7) | 3 |
| *: since Kitti disparity map ground truth masked out person and bike categories, we can only compare against car category. | |

According to the abs disparity error histogram, 85.8% of predicted object disparity are within 3 pixesl of error and 98.9% of the predicted object disparity are within 5 pixels of error. This is achieved by training only on a small dataset of 514 stereo images from scratch. We believe by enlarging the training dataset the mean/max abs disparity error can be further reduced.

# 5. Edge Computing Devices Benchmarking

Since we are using squeeznet as our base net, the model size is relatively small, which is about 5.7 MB at 32FP precision and 1.7MB at INT8 precision. This enables us running the model inference at a resource limited edge computing device such as Nvidia Jetson (Nano, AGX) as well as Rockchip 3399Pro, 3566, etc.

Table 3 lists 1 desktop graphic card (for reference purpose) and 4 edge computing devices configuration and inference + NMS performance of running the model in different precision with input stereo image resolution of 640x640.

| Table 3: Performance benchmark at 640x640 | | | | | |
|---|---|---|---|---|---|
| Device | Spec | TOPS rating | model precision size | Inference only | Inference + NMS |
| 1080Ti | 3584 cores i7-5820K | | Pytorch FP32 5.7MB | 10.3 ms/frame | 18.8 ms/frame |
| Nano | 128 cores 4x A57 | 0.47 | Pytorch FP32 5.7MB | 40.2 ms/frame | 320.5 ms/frame |
| Agx | 512 cores 8x Arm.v8.2 | 32 | Pytorch FP32 5.7MB | 24.4 ms/frame | 39.7 ms/frame |
| RK3399 Pro | 2x A72, 4x A53, NPU | 3 | Python* FP16 2.97MB | 238.9 ms/frame | 367 ms/frame |
| RK3399 Pro | 2x A72, 4x A53, NPU | 3 | Python* INT8 1.58MB | 25.9 ms/frame | 84.4 ms/frame |
| RK3566 | 4x A55, NPU | 0.8 | Python* INT8 1.7MB | 97.7 ms/frame | 429.3 ms/frame |
| RK3566 | 4x A55, NPU | 0.8 | C** INT8 1.7MB | 93.7 ms/frame | 115.8 ms/frame |
| *:RKNN Python API; **:RKNN C API | | | | | |

If we can reduce the far distance detection precision and recall rate, then we can reduce the input image resolution down to 320x320 and the inference + NMS speed can be further improved.

| Table 4: OD-SSD test results at 320x320 | |
|---|---|
| Total test stereo images | 86 |
| Precision (car*) | 85.2% |
| Recall (car*) | 52.4% |
| Mean abs obj disparity error | 2.43 pixels |
| Max abs obj disparity error | 8.89 pixels |
| Abs obj disparity error histogram (pixel) | Number of objects |
| [0, 1) | 47 |
| [1, 2) | 57 |
| [2, 3) | 25 |
| [3, 4) | 25 |
| [4, 5) | 10 |
| [5, 6) | 10 |
| [6, 7) | 7 |
| [7,8) | 3 |
| [8,9) | 5 |
| *: since Kitti disparity map ground truth masked out person and bike categories, we can only compare against car category. | |

It looks like the precision and recall rate dropped significantly when the input resolution was reduced from 640x640 to 320x320. However, if we look at the sample stereo image test result in Figure 8 (compared with Figure 6), we will see that the majority of detection precision and recall rate drops are due to the far away object lost. The close-by object detection and object disparity accuracy are still intact.

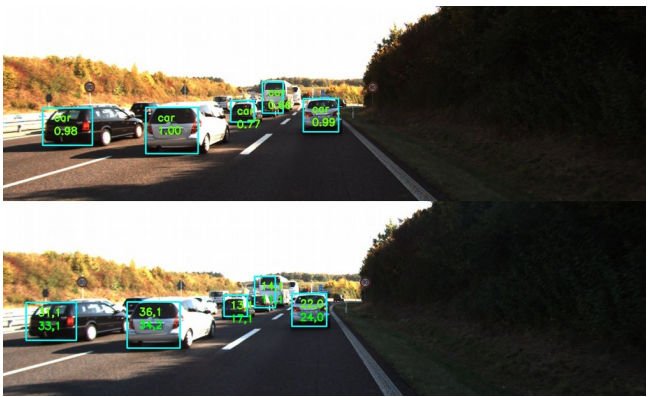

*Figure 8: OD-SSD test result at 320x320 input resolution*

Table 5 lists the performance benchmark at 320x320 input resolution. As expected, the low end edge devices (such as Nano, RK3566 and RK3399Pro) have significant performance improvement.

| Table 5: Performance benchmark at 320x320 | | | | | |
|---|---|---|---|---|---|
| Device | Spec | TOPS rating | model precision size | Inference only | Inference + NMS |
| 1080Ti | 3584 cores i7-5820K | | Pytorch FP32 5.7MB | 11.0 ms/frame | 15.4 ms/frame |
| Nano | 128 cores 4x A57 | 0.47 | Pytorch FP32 5.7MB | 35.0 ms/frame | 95.0 ms/frame |
| Agx | 512 cores 8x Arm.v8.2 | 32 | Pytorch FP32 5.7MB | 23.7 ms/frame | 33.8 ms/frame |
| RK3399 Pro | 2x A72, 4x A53, NPU | 3 | Python* FP16 2.97MB | 81.4 ms/frame | 153.6 ms/frame |
| RK3399 Pro | 2x A72, 4x A53, NPU | 3 | Python* INT8 1.58MB | 8.9 ms/frame | 52.0 ms/frame |
| RK3566 | 4x A55, NPU | 0.8 | Python* INT8 1.6MB | 25.4 ms/frame | 173.4 ms/frame |
| RK3566 | 4x A55, NPU | 0.8 | C** INT8 1.6MB | 25.8 ms/frame | 29.2 ms/frame |
| *:RKNN Python API; **:RKNN C API | | | | | |

# 6. Object Disparity-SSD Robustness and Stereo-Parameter-Agnostic Test

In this section, we would like to demonstrate the robustness and stereo-parameter-agnostic of OD-SSD by using a mixed dataset captured by several uncalibrated stereo camera systems and a calibrated stereo camera system for training and testing.

We used microAI.systems iVidar stereo camera system[11] for capturing two uncalibrated stereo image datasets. By "uncalibrated stereo images", we mean that the stereo images captured have not run through camera undistortion transformation neither performed stereo calibration for epipolar line correction.

As shown in Figure 9, the iVidar has two stereo subsystems: S1 consists of cameras 102 and 104; S2 consists of cameras 106 and 108. The S1 has a baseline of 10 cm with FOV of 150º. The S2 has a baseline of 25 cm with FOV of 50º.

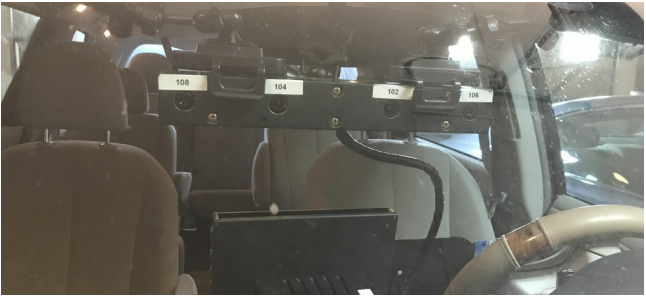

*Figure 9: microAI.systems iVidar*

And we used two off-the-shelf DashCams as shown in Figure 10 to capture a third uncalibrated stereo image dataset.

Then we mixed 4 different stereo images captured by 4 following different stereo systems as listed in Table 6 to form a mixed stereo image dataset:

| Table 6: mixed stereo image dataset sources | | | | |
|---|---|---|---|---|
| Dataset | Stereo System | Baseline | FOV | Stereo calibrated |
| S1 | iVidar S1 | 10 cm | 150º | No |
| S2 | iVidar S2 | 25 cm | 50º | No |
| DashCam | DashCam | 20 cm | 170º | No |
| Kitti | Kitti | 54 cm | 71.4º~35.7º | Yes |

The training dataset consists of 882 images from above mixed stereo image dataset. The testing data set has 22 mixed stereo images also captured by the above 4 different stereo systems.

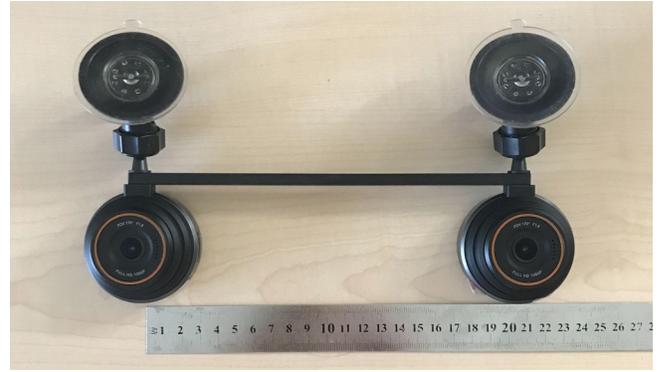

*Figure 10: DashCam Stereo System with 20 cm baseline*

The test results are summarized in Table 7. These results may suggest that with adequate size of mixed training dataset an OD-SSD might be agnostic to stereo system parameters such as a baseline, FOV, lens distortion, even left/right camera epipolar line misalignment.

| Table 7: OD-SSD test results at 640x640 with mixed dataset | |
|---|---|
| Total test stereo images | 22 |
| Precision (car*) | 88% |
| Recall (car*) | 89.3% |
| Mean abs obj disparity error | 2.86 pixels |
| Max abs obj disparity error | 9.4 pixels |
| Abs obj disparity error histogram (pixel) | Number of objects |
| [0, 1) | 12 |
| [1, 2) | 14 |
| [2, 3) | 4 |
| [3, 4) | 6 |
| [4, 5) | 2 |
| [5, 6) | 1 |
| [6, 7) | 8 |
| [7,8) | 0 |
| [8,9) | 1 |
| [9,10) | 1 |
| *: since Kitti disparity map ground truth masked out person and bike categories, we can only compare against car category. | |

A test image captured by S2 on a rainy night is shown in Figure 11. The image resolution captured by S2 is 1920x1080. As we can see the object disparity accuracy is relatively high (predicted dx=144 vs. ground truth dx=145) for the object in the center field of view. Since we use an "uncalibrated stereo image", we can see that even with the ground truth dy=10 (that is, the left image epipolar line is offset by 10 pixels from right image epipolar line), the object disparity prediction error is still within 1 pixel.

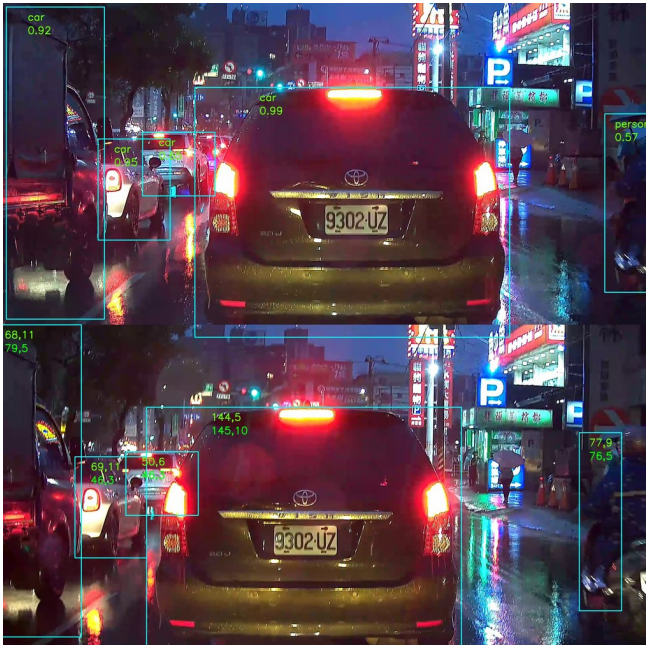

*Figure 11: Test Image captured by S2*

To demonstrate that the OD-SSD is indeed capable of "computing" disparity, we constructed a "fake" test stereo image by a photo editor to move a car at the center by 98 pixels of disparity in the right image and kept the left image intact. The original real disparity of the car in the right image is 24 pixels as shown in Figure 12. The OD-SSD actually detects the car in the "fake" test stereo image with "correct" disparity of 98 pixels as shown in Figure 13.

This test may imply that the OD-SSD is not only able to detect the same object in both left and right images but also capable of "computing" the correct disparity of the detected object from left and right images.

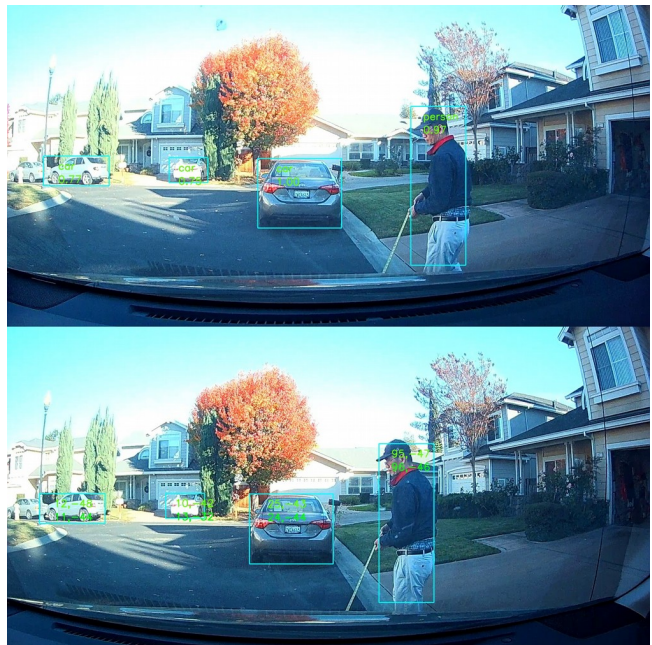

*Figure 12: Original real stereo images captured by DashCam system with baseline of 20 cm.*

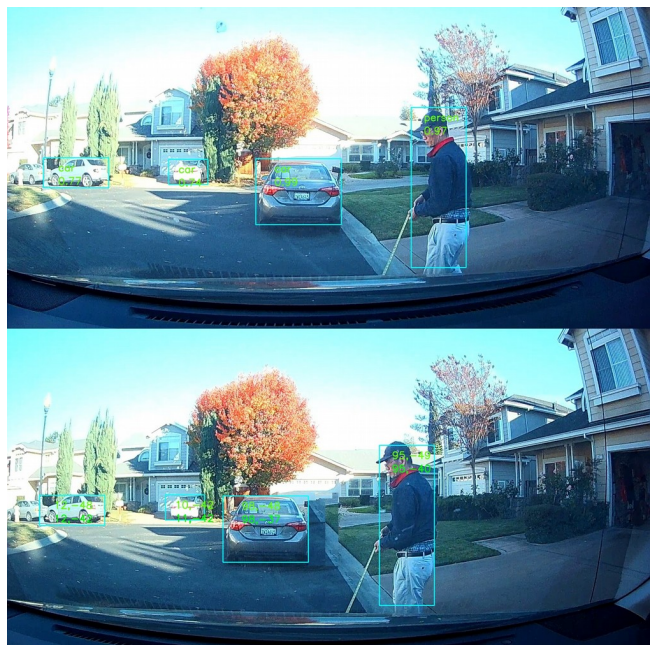

*Figure 13: a "fake" test image by editing right image to move the car in the center by 98 pixel of disparity*

## 7. Conclusion

A new approach for solving a 3D object distance detection by detecting object disparity directly without going through a dense pixel disparity computation was proposed.

An example squeezenet Object Disparity-SSD (OD-SSD) was constructed. It was trained on 514 Kitti stereo images from scratch and reached 98.9% of the predicted object disparity within 5 pixels of error. The model size of OD-SSD is only 5.7 MB at 32FP precision and 1.7MB at INT8 precision.

The performance benchmark on 4 different edge devices showed the potential of running the OD-SSD in real time in a low-end edge device such as Jetson Nano, Rockchip 3399Pro, or even Rockchip 3566 at input resolution of 320x320.

Further training and testing results with a mixed image dataset captured by several different stereo systems may suggest that an OD-SSD might be agnostic to stereo system parameters such as a baseline, FOV, lens distortion, even left/right camera epipolar line misalignment.

We believe by enlarging the training dataset the mean/max abs disparity error can be further reduced.